\documentclass[runningheads]{llncs}
\usepackage{graphicx}
\usepackage{amsmath}
\usepackage{caption}
\usepackage{subcaption}
\usepackage{booktabs}
\usepackage{array} 
\usepackage{booktabs}  
\usepackage{xcolor}    
\usepackage{colortbl}  
\usepackage{amssymb}   
\usepackage{booktabs}  
\usepackage{xcolor}    
\usepackage{colortbl}  
\usepackage{amssymb}   
\usepackage{graphicx}  
\usepackage{array}     
\usepackage[T1]{fontenc}
\usepackage[table]{xcolor}
\usepackage{booktabs}
\usepackage{graphicx}
\usepackage{caption}
\usepackage{orcidlink}
\usepackage{pifont} 
\usepackage{multirow}

\usepackage{graphicx,verbatim}
\begin{document}
\title{MedAgentSim: Self-Evolving Multi-Agent Simulations for Realistic Clinical Interactions}
\titlerunning{Multi-Agent Simulations for Realistic Clinical Interactions}

\author{Mohammad Almansoori$^{*, \dagger}$\orcidlink{0009-0007-8590-0428} \and
Komal Kumar$^{*}$\orcidlink{0000-0002-9165-9390} \and
Hisham Cholakkal\orcidlink{0000-0002-8230-9065}}

%
\authorrunning{M. Almansoori et al.}

\institute{Mohamed bin Zayed University of Artificial Intelligence, Abu Dhabi, UAE\\
\email{\{mohammad.almansoori, komal.kumar, hisham.cholakkal\}@mbzuai.ac.ae}\\
$^{*}$Equal contribution \quad $^{\dagger}$Corresponding author}

\maketitle              
\begin{abstract}
In this work, we introduce MedAgentSim, an open-source simulated clinical environment with doctor, patient, and measurement agents designed to evaluate and enhance LLM performance in dynamic diagnostic settings. Unlike prior approaches, our framework requires doctor agents to actively engage with patients through multi-turn conversations, requesting relevant medical examinations (e.g., temperature, blood pressure, ECG) and imaging results (e.g., MRI, X-ray) from a measurement agent to mimic the real-world diagnostic process. 
Additionally, we incorporate self improvement mechanisms that allow models to iteratively refine their diagnostic strategies. We enhance LLM performance in our simulated setting by integrating multi-agent discussions, chain-of-thought reasoning, and experience-based knowledge retrieval, facilitating progressive learning as doctor agents interact with more patients. We also introduce an evaluation benchmark for assessing the LLM’s ability to engage in dynamic, context-aware diagnostic interactions. While MedAgentSim is fully automated, it also supports a user-controlled mode, enabling human interaction with either the doctor or patient agent. Comprehensive evaluations in various simulated diagnostic scenarios demonstrate the effectiveness of our approach. Our codebase, simulation environment, and benchmark datasets are publicly available on the project page: \url{https://medagentsim.netlify.app/}.

\end{abstract}

\section{Introduction}
Advancements in Large Language Models (LLMs) and Vision-Language Models (VLMs) have shown promising capabilities across various medical tasks, achieving human-level performance on several medical benchmarks \cite{nori2023can}.  
These models have demonstrated the ability to encode clinical knowledge \cite{singhal2023large,vaid2023using}, retrieve relevant medical literature \cite{xiong2024benchmarking}, and achieve high accuracy in single-turn medical question-answering tasks \cite{chen2023meditron,lievin2024can,nori2023can,wu2024pmc}. However, current medical LLM assessments   often rely on static evaluation benchmarks, where models are provided with complete patient information and tasked with answering predefined questions, sometimes with multiple-choice options \cite{jin2021disease}. These assessments often  fail to capture the complexity of real-world doctor-patient interactions, where diagnosis is not a single-step process but a dynamic, multi-turn dialogue.  Such multi-turn doctor-patient interactions are important in clinical scenarios, as patients often struggle to describe their symptoms accurately due to limited medical knowledge, ambiguous perceptions, or communication barriers  \cite{meyer2021patient}. Consequently, physicians play an active role in structuring these interactions, posing clarifying questions, and refining their assessments as new information emerges \cite{zhong2022hierarchical}. 



\begin{figure}[ht]
    \centering
    \begin{subfigure}{0.69\linewidth}
        \centering
        \includegraphics[width=\linewidth]{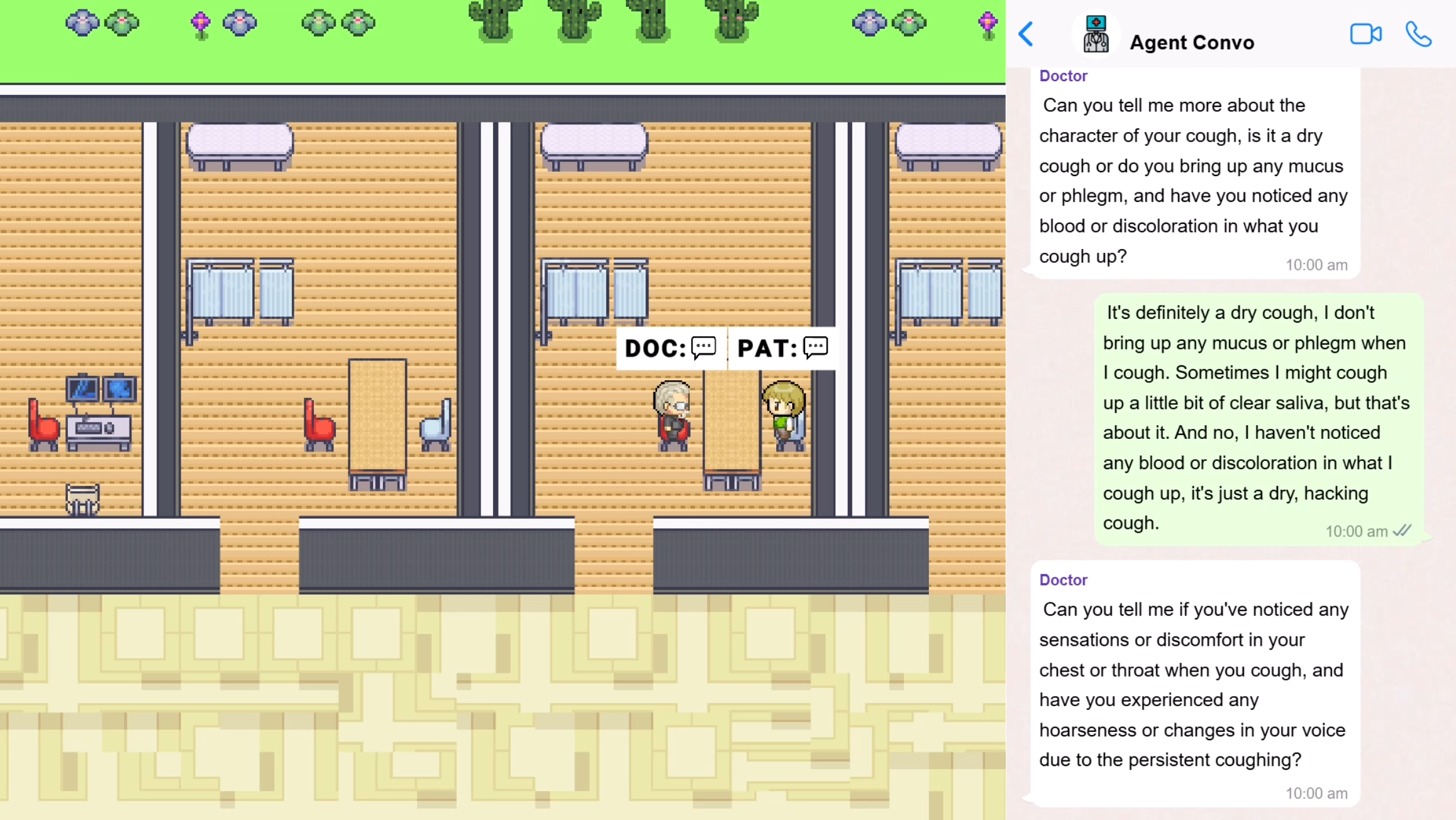}
        \caption{ Screenshot of our simulation environment showing  doctor-patient interaction phase, where the doctor agent gathers  clinical information via multi-turn conversation.}
        \label{fig:conversation}
    \end{subfigure}
    \hfill
    \begin{subfigure}{0.29\linewidth}
        \centering
        \includegraphics[width=0.9\linewidth, height=4.75cm]{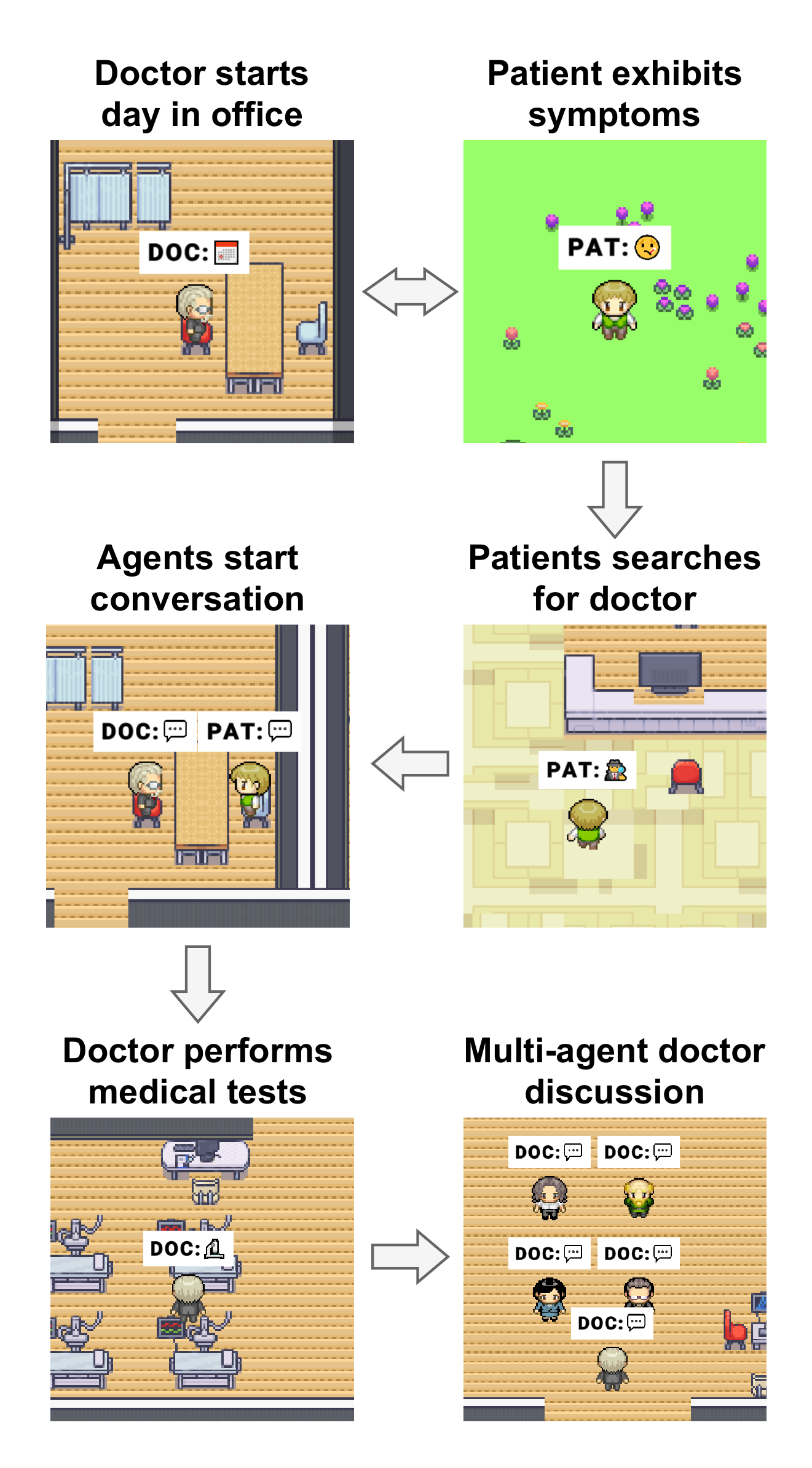}
        \caption{ The sequential progression of the simulation and events at each stage.}
        \label{fig:sim}
    \end{subfigure}
    \caption{Interactive clinical simulations in our MedAgentSim (best viewed when zoomed in).}
    \label{fig:combined}
\end{figure}

Despite the aforementioned clinical significance, recent studies have highlighted that LLMs struggle in realistic clinical scenarios where they are not provided with all relevant information upfront \cite{fan2025ai,schmidgall2024agentclinic}. 
Instead, they \cite{fan2025ai,schmidgall2024agentclinic} shared  only limited initial knowledge about the patient to the LLM and the LLMs are required to engage in a dynamic diagnostic process, systematically refining their understanding through patient dialogue. However, approaches such as AI Hospital \cite{fan2025ai} only introduced evaluation benchmarks, without  enhancing LLMs for multi-turn interactions. Additionally, they relied on chat-based textual interaction simulations, where LLMs were not required to navigate complex environments or interact with medical tools. 

Recently, LLM-driven game simulations were introduced in \cite{li2024agent} for clinical settings, where closed-source AI agents based on OpenAI GPT-4o \cite{openai2024chatgpt4o} were assigned roles such as doctors and patients \cite{li2024agent}. These simulations were effective in capturing several aspects of real-world clinical complexity by requiring agents to navigate environments, interact with objects, and engage dynamically in decision-making. Additionally, these studies \cite{du2024llms,li2024agent} incorporated memory-replay techniques to enhance agent performance. However, these approaches deviate from real-world clinical practice, as doctor agents are provided with\textit{a pre-compiled, complete medical report of the patient}, rather than doctor agents actively gathering patient information through interactive consultations. Furthermore, these simulations lack the ability to incorporate  medical image-based diagnostic resources such as X-Rays and CT scans, which are critical in real medical decision-making. In addition to relying on closed-source LLMs like GPT-4o, many of these systems remain closed-source, limiting access to their data, code, and models, which hinders reproducibility and further research.

To address the limitations of existing methods, \textbf{\textit{we introduce MedAgentSim, an open-source, simulated hospital environment }} designed to  \textit{evaluate and enhance} LLM performance in dynamic diagnostic settings. Unlike prior approaches, our framework, illustrated in Figure \ref{fig:conversation}, requires doctor agents to \textit{actively engage with patients through multi-turn conversations}, \textit{prompting medical examinations} to capture vital signs such as temperature, blood pressure, and electrocardiogram (ECG), \textit{and requesting imaging results} (e.g., MRI, X-Ray) prior to making a diagnosis. Furthermore, we incorporate \textit{self-improvement mechanisms}, allowing the models to iteratively refine their diagnostic strategies over time. We also \textit{introduce an evaluation benchmark} designed to bridge the gap between static evaluations and real-world medical reasoning by assessing the LLM agent's ability to engage in dynamic, context-aware diagnostic interactions, bringing it one step closer to practical clinical applications. 

The \textbf{\textit{key contributions}} of our method are summarized as below:
\begin{enumerate} \item A game-based hospital simulation built with open-source LLMs \cite{meta2024llama33,mistral2025small3}, where LLM-powered doctor and patient agents interact in a realistic diagnostic setting.  The system is fully automated and it also supports a user-controlled mode, allowing a human to take control of either the doctor or patient agent for real-time interaction with the AI counterpart. 
    \item A multi-agent LLM framework for realistic doctor-patient dialogue, where the doctor starts with no prior knowledge of the patient's condition and need to  ask questions for gathering relevant patient information. Test results are only provided if the doctor specifically requests the necessary tests, ensuring a process that closely mirrors real-world clinical consultations.
    \item A multi-agent diagnostic pipeline that improves baseline LLM performance by incorporating self-improvement mechanisms, including multi-agent discussion, chain-of-thought (COT) \cite{wei2022chain} reasoning, and experience-based knowledge retrieval. The system enables progressive learning, where doctor agents refine their diagnostic capabilities as they interact with more patients.
\end{enumerate}

\section{Methodology: MedAgentSim}
\begin{figure*}[!h]
    \centering
    \begin{subfigure}{0.74\linewidth}
        \centering
        \includegraphics[width=\linewidth]{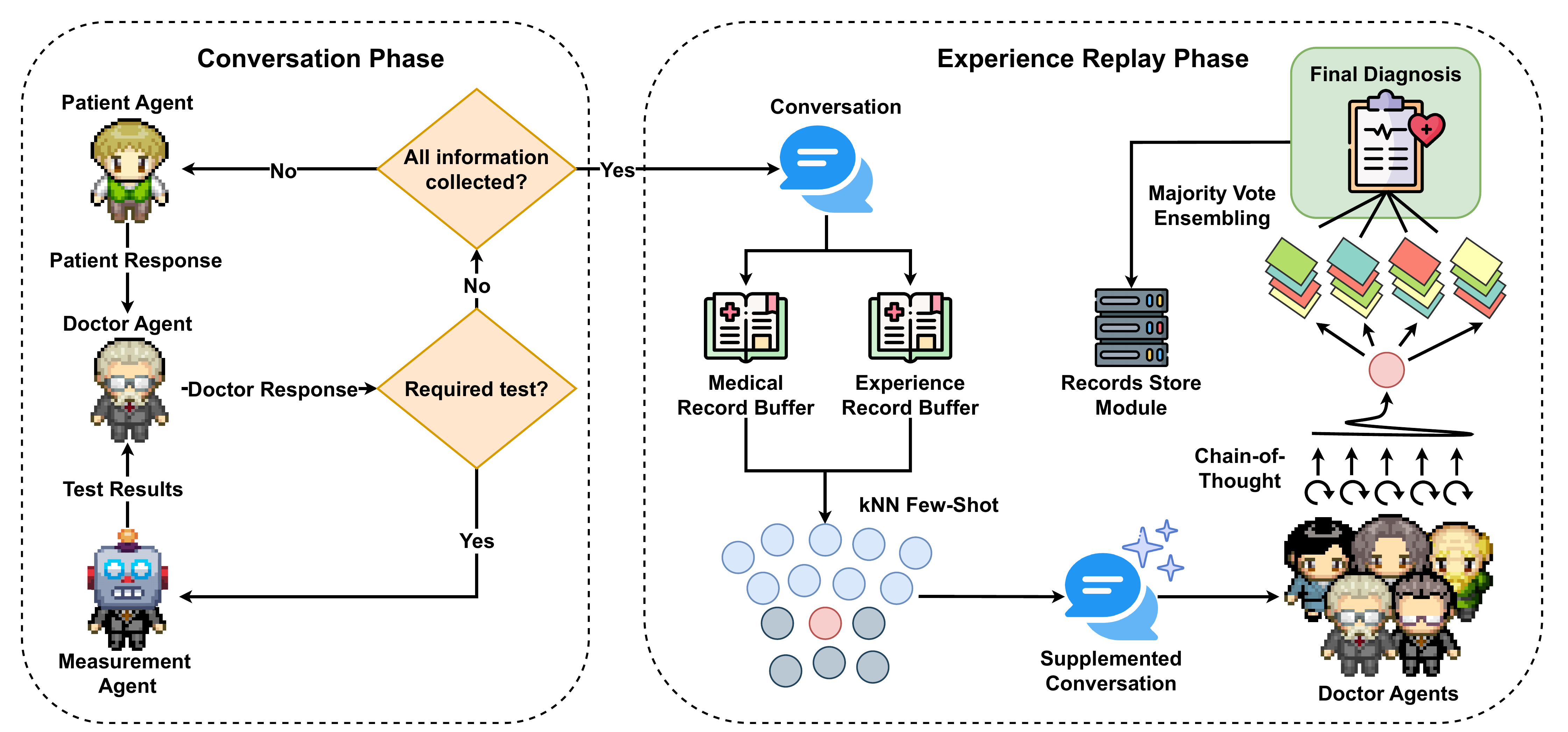}
        \caption{In \textit{Conversation Phase}, the doctor and patient agents engage in an interactive dialogue, allowing the doctor to gather vital information and request necessary diagnostic tests, such as blood tests and X-Rays, from the measurement agent. As results are provided, the conversation continues until the doctor has sufficient information. Once ready to diagnose, the process transitions to \textit{Experience Replay Phase}. Here, past doctor-patient interactions are analyzed through memory buffers, retrieving relevant cases as few-shot examples to enrich the current dialogue. A team of doctor agents then evaluates this enhanced conversation using \textit{COT reasoning} and \textit{majority-vote ensembling} to reach a consensus, producing a well-informed diagnosis.}
        \label{fig:architecture}
    \end{subfigure}
    \hfill
    \begin{subfigure}{0.24\linewidth}
        \centering
        \includegraphics[width=0.9\linewidth, height=4.5cm]{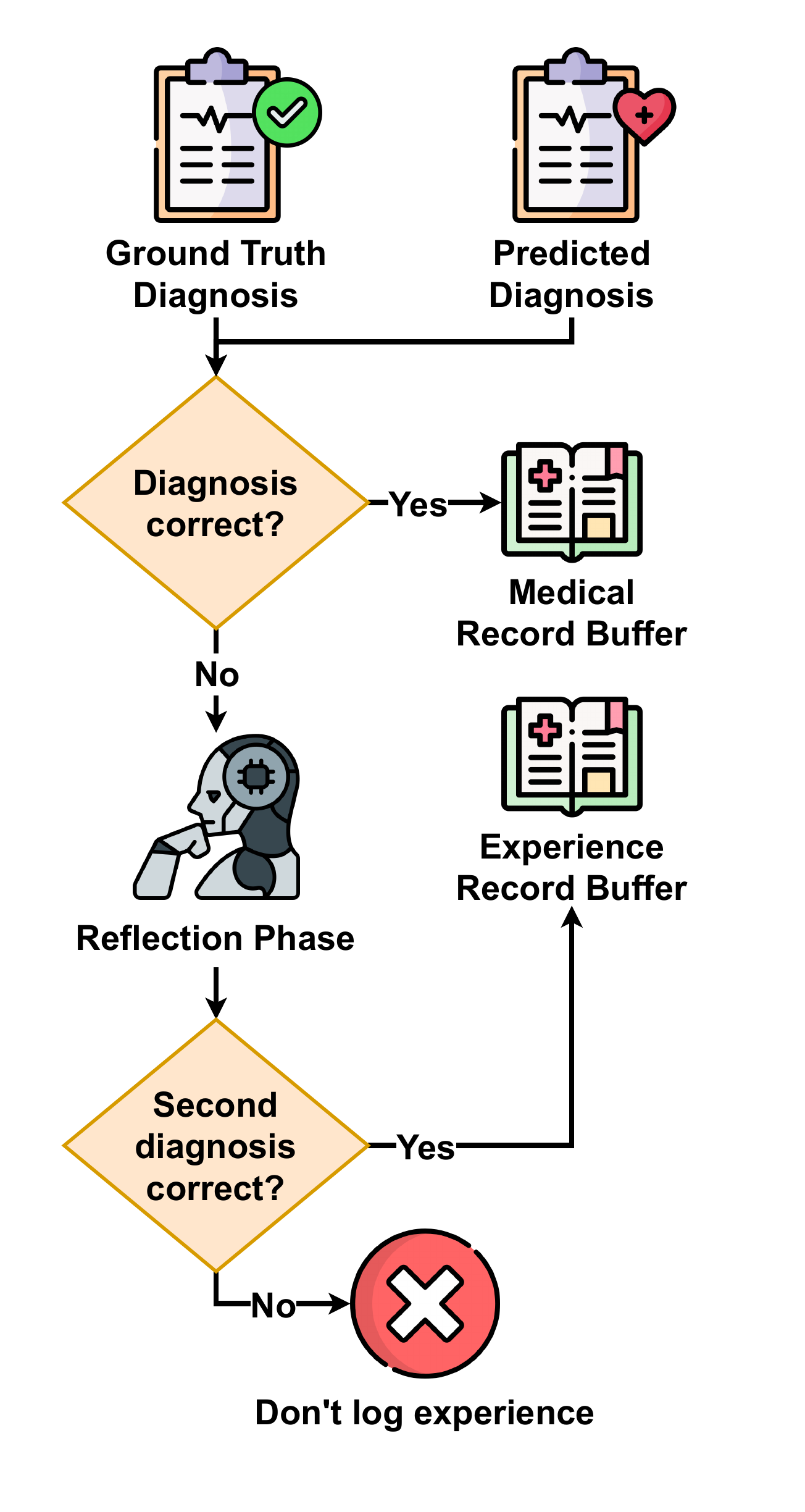}
        \caption{The record storing module progressively maintains a \textit{medical records buffer} for storing correct diagnoses and an \textit{experience records buffer} for tracking cases where initial misdiagnoses were later corrected upon reflection.}
        \label{fig:reflect}
    \end{subfigure}
    \caption{\textbf{(a)} Overview  of the proposed MedAgentSim comprising   \textit{Conversation} and \textit{Experience Replay} phases.  \textbf{(b)} Our records store module for progressive learning.}
    \label{fig:combined_pipeline}
\end{figure*}

Figure \ref{fig:combined_pipeline} shows an overview of the proposed MedAgentSim comprising  two key phases. At first, in the  \textit{Conversation Phase},  agents actively gather all relevant patient information necessary for diagnosis. Then, in the \textit{Experience Replay Phase},  correctly diagnosed cases are stored as memory for future retrieval and learning. Next, we introduce our overall simulation architecture.

\noindent\textbf{Simulation Environment.} The proposed hospital simulation environment builds upon Generative Agents \cite{park2023generative}, transforming it into an interactive healthcare setting where autonomous virtual characters, commonly referred to as non-playable characters (NPCs), simulate real-world hospital dynamics. These NPCs, powered by an LLM, can move freely, initiate conversations, and interact with medical equipment, making real-time decisions based on the unfolding scenario.

\noindent\textbf{(a) Agent Roles.} The simulation consists of three core agent types: the \textit{patient agent}, the \textit{doctor agent}, and the \textit{measurement agent}. The patient experiences symptoms and seeks medical attention from the doctor, who is responsible for diagnosing and treating conditions. The measurement agent provides diagnostic test results but only when explicitly requested, requiring the doctor to actively gather information rather than receiving all patient data upfront. Figure \ref{fig:sim} showcases a sample scenario, demonstrating how agents navigate the environment and engage in clinical workflows. This baseline framework is referred to as Multi-Agent Clinic \cite{schmidgall2024agentclinic}.
 
\noindent\textbf{(b) Agent Interaction Modes.} Both the doctor and patient agents can function in one of three distinct modes, determining how they generate and process information during interactions. In \textit{Generation Mode}, the patient agent autonomously creates a case, generating illnesses, symptoms, and test results, which are internally stored. The doctor agent must actively extract relevant details through questioning. In \textit{Dataset Mode}, patient responses are derived from a predefined dataset, ensuring consistency with structured medical knowledge, while the doctor agent follows the same interactive probing process. Finally, in \textit{Control Mode}, a human user can assume control of either the doctor or patient, enabling real-time interactions with the AI-driven counterpart. This mode facilitates testing and supports potential real-world deployment, where real patients could engage with an AI-powered doctor or vice-versa.

\noindent\textbf{Memory and Self-Improvement.} Doctor-patient consultations take place through natural language interactions, where the doctor questions the patient, infers possible conditions, and orders tests. If a medical test is not requested, its results remain unavailable, mirroring real-world diagnostic constraints. Once the doctor is ready to make a diagnosis, the conversation undergoes a experience replay phase, refining the model’s decision-making over time.

\noindent\textbf{(a) Records Buffer.} To enable progressive learning, the system maintains a record storage and retrieval mechanism that captures both successful and corrected diagnoses. It consists of two dynamically expanding libraries: the \textit{Medical Records Buffer}, which stores correctly diagnosed cases, and the \textit{Experience Records Buffer}, which retains misdiagnosed cases that were later corrected through reflection. During a new consultation, the system uses k-nearest neighbors (KNN) to retrieve relevant past cases. The Medical Records Buffer provides full conversations and diagnoses, while the Experience Records Buffer extracts key insights from the reflection process. This approach builds on prior experiences, as studies show that LLMs benefit from failure-driven learning \cite{yang2023failures}.

\noindent\textbf{(b) COT and Ensembling.} The retrieved information is then incorporated into the consultation, enriching the doctor’s contextual understanding. A multi-agent system processes the updated input, where multiple doctor agents independently assess the case and propose diagnoses. These assessments are aggregated and refined using COT \cite{wei2022chain} reasoning and majority-vote ensembling \cite{nori2023can}, producing a final diagnosis.

\noindent\textbf{(c) Records Storage.} Once finalized, the system converts all case data, including conversation history, diagnosis, medical images, and lab results, into CLIP \cite{radford2021learning} embeddings. Correct diagnosis embeddings are added to the Medical Records Buffer, while incorrect cases undergo a reflection phase where the doctor analyzes the mistake before making a second attempt. If the revised diagnosis is correct, only the CLIP-embedded reflection insights are stored in the Experience Records Buffer; otherwise, the case is discarded to ensure learning is based on meaningful examples. Figure \ref{fig:reflect} illustrates the full reflection and storage process.

\section{Experiments}
\noindent\textbf{Experimental Details.}
We conducted extensive experiments to evaluate the effectiveness of MedAgentSim in a real-world doctor-agent setting. Our study leveraged a diverse set of both open-source models available on Hugging Face \cite{hf} and proprietary models, tested across three primary benchmarks: NEJM \cite{schmidgall2024agentclinic}, MedQA \cite{jin2021disease}, and MIMIC-IV \cite{johnson2023mimic}.
For VLM tasks, we utilized the NEJM dataset, which includes 15 complex real-world cases along with an extended set, NEJM Extended, of 120 additional cases. MedQA comprises 106 simulated diagnostic scenarios, while its extended variant, MedQA Extended, contains 214 distinct cases. Additionally, MIMIC-IV features 288 clinical cases, providing a diverse set of real-world medical interactions.
As these datasets are primarily formatted for QA tasks, they are not directly compatible with our simulation pipeline. To address this, we preprocess the data using GPT-4o, converting it into a structured JSON format, where the doctor, patient information, and test results are assigned to the doctor agent, patient agent, and measurement agent, respectively. Model accuracy is evaluated using a \textit{binary true/false metric} for the final diagnosis, with an LLM serving as the evaluator to account for variability in generated responses. Both the dataset conversion process and accuracy logs were manually reviewed to ensure reliability.

All models were deployed using vLLM \cite{kwon2023efficientmemorymanagementlarge} on a 4×48 GB NVIDIA RTX A6000 setup. For vision-language tasks, we integrated QwenVL \cite{qwen25-72b}, for the Qwen family of models, and LLaVA 1.5 \cite{liu2023visual} for the remaining models, with LLaVA demonstrating strong performance in medical image interpretation, particularly in generating descriptive reports for X-Rays, MRIs, and other imaging modalities. The visual game simulation was developed using Phaser, a web-based game engine \cite{Phaser_Studio_2018}, with the map designed in Tiled, a 2D level editor \cite{Tiled_2019}. Game assets were sourced from Generative Agents \cite{park2023generative}.

\noindent\textbf{Results and Analysis.}
Table \ref{tab:multi_agent_performance} compares the performance of the baseline Multi-Agent Clinic and our proposed MedAgentSim across key medical benchmarks, covering both language-based and vision-based tasks. MedAgentSim integrates LLaVA 1.5-Mistral, a multi-modal model combining visual encoding with large language models.

The results show that MedAgentSim significantly outperforms the baseline across all benchmarks, particularly in multi-modal tasks. In the NEJM benchmark, MedAgentSim achieves 26.7\% with LLaMA 3.3, a substantial improvement over the baseline Multi-Agent Clinic, where models struggle to exceed 20.0\%. This gap widens in NEJM Extended, where MedAgentSim reaches 28.3\% with LLaMA 3.3, surpassing the best baseline performance of 24.2\%. These findings indicate that MedAgentSim is better equipped to interpret medical images and generate accurate clinical insights.

For language-based reasoning, MedAgentSim consistently demonstrates superior performance. In MedQA, it achieves 70.8\% with LLaMA 3.3, while the best-performing baseline model records 62.3\%. Similarly, in MedQA Extended, MedAgentSim attains 72.0\%, a notable increase over the 63.6\% baseline. The most significant performance boost is observed in MIMIC-IV, where MedAgentSim reaches 79.5\%, far exceeding the highest baseline score of 42.7\%. 

In addition to automated benchmarks, a preliminary human study was performed in which one agent (doctor, patient, or measurement) was replaced with a real human. Doctors viewing the clips failed to identify the human in 62.5\% of cases, indicating high behavioral realism. Full results are on the project page\footnote{\url{https://medagentsim.netlify.app}}.

\begin{table}[htbp]
    \centering
    \caption{Performance of Multi-Agent Clinic (Basic) and MedAgentSim (Our) models across medical benchmarks. We used diverse LLMs including closed source. For visual language tasks, we use LLava 1.5 \cite{liu2023visual} for visual encoding.}
    \label{tab:multi_agent_performance}
    \rowcolors{2}{gray!15}{white}
    \renewcommand{\arraystretch}{1.2} 
    \resizebox{0.7\textwidth}{!}{ 
    \begin{tabular}{llc ccccc}
        \toprule
        \rowcolor{gray!30} & Baseline & Size/Type & NEJM & NEJM Ext. & MedQA & MedQA Ext. & MIMIC-IV \\
        \midrule
        \multicolumn{8}{c}{\textbf{Multi-Agent Clinic}} \\  
        \midrule
        & Claude \cite{anthropic2024claude35} & 3.5 & — & — & 62.3 & 63.6 & 42.7 \\
        & ChatGPT \cite{openai2024chatgpt4o} & 4o & 26.7 & 25.8 & 52.8 & 52.3 & 34.4 \\
        & ChatGPT \cite{openai2023chatgpt4} & 4 & 13.3 & 19.2 & 35.8 & 33.2 & 24.7 \\
        & ChatGPT \cite{openai2022chatgpt3.5} & 3.5 & — & — & 36.8 & 34.6 & 27.8 \\
        & LLaMA 3.3 \cite{meta2024llama33} & 70B & 20.0 & 24.2 & 54.7 & 53.3 & 36.8 \\
        & LLaMA 3 \cite{dubey2024llama3} & 70B & 6.7 & 5.0 & 19.8 & 17.3 & 13.9 \\
        & LLaMA 2 \cite{touvron2023llama2} & 70B & — & — & 4.7 & 2.8 & 8.3 \\
        & Mixtral \cite{jiang2024mixtral8x7b} & 8×7B & 6.7 & 2.5 & 37.7 & 39.3 & 30.2 \\
        & Mistral \cite{mistral2025small3} & 24B & 6.7 & 3.3 & 45.3 & 41.1 & 21.9 \\
        & Qwen2 \cite{Qwen2-VL} & VL-7B & 0.0 & 1.7 & 20.8 & 16.8 & 25.7 \\
        & Qwen2.5 \cite{qwen25-72b} & 72B & 0.0 & 2.5 & 38.7 & 41.6 & 21.2 \\
        \midrule
        \multicolumn{8}{c}{\textbf{MedAgentSim (Ours)}} \\
        \midrule
        & ChatGPT \cite{openai2024chatgpt4o} & 4o & 26.7 & 27.5 & 66.0 & 67.8 & 75.3 \\
        & LLaMA 3.3 \cite{meta2024llama33} & 70B & 26.7 & 28.3 & 70.8 & 72.0 & 79.5 \\
        & Mistral \cite{mistral2025small3} & 24B & 13.3 & 9.2 & 53.8 & 49.5 & 56.6 \\
        & Qwen2 \cite{Qwen2-VL} & VL-7B & 6.7 & 4.2 & 31.3 & 29.2 & 38.2 \\
        & Qwen2.5 \cite{qwen25-72b} & 72B & 6.7 & 4.2 & 55.7 & 57.5 & 66.0 \\
        \bottomrule
    \end{tabular}
    }
\end{table}



\subsection{Ablation Study}
\textbf{Impact of MedAgentSim Strategies.}
Table~\ref{tab:performance} summarizes the impact of adding incremental reasoning strategies on model accuracy. The integration of measurement, memory augmentation, COT \cite{wei2022chain} reasoning, and ensembling progressively improves diagnostic performance. Notably, the LLaMa 3.3 70B model benefits significantly from memory and COT strategies, achieving a final accuracy boost of 16.1\%.
\begin{table}[h]
  \centering
  \rowcolors{2}{gray!15}{white}
  \renewcommand{\arraystretch}{1.2} 
  \caption{Incremental improvements in model accuracy as measurement, memory, COT \cite{wei2022chain}, and ensembling techniques are added.}
  \label{tab:performance}
  \resizebox{0.5\linewidth}{!}{ 
  \begin{tabular}{l c | l c}
    \toprule
    \rowcolor{gray!30} \textbf{Mistral 24B} & \textbf{Accuracy} & \textbf{LLaMa 3.3 70B} & \textbf{Accuracy} \\
    \midrule
    Baseline            & 45.3\%  & Baseline          & 54.7\% \\
    + Measurement       & 47.2\%  & + Measurement     & 59.4\% \\
    + Memory           & 51.9\%  & + Memory          & 65.1\% \\
    + COT              & 52.8\%  & + COT             & 68.9\% \\
    + Ensembling       & 53.8\%  & + Ensembling      & 70.8\% \\
    \bottomrule
  \end{tabular}}
\end{table}

\begin{table}[h]
  \centering
  \rowcolors{2}{gray!15}{white}
  \renewcommand{\arraystretch}{1.2}
  \caption{Doctors’ ability to identify the real human in each video. \ding{51} = Correct guess, \ding{55} = Incorrect guess.}
  \label{tab:human-identification-final}
  \resizebox{\linewidth}{!}{
  \begin{tabular}{l p{2.2cm} p{2.2cm} p{3.2cm} c}
    \toprule
    \rowcolor{gray!30}
    \textbf{Video ID} & \textbf{Doctor ID} & \textbf{Guess} & \textbf{Ground Truth} & \textbf{Accuracy} \\
    \midrule
    VIDEO\_01 & Doctor 1 & None         & None (All AI)           & \ding{51} \\
              & Doctor 2 & Patient      &                         & \ding{55} \\
    VIDEO\_02 & Doctor 1 & Patient      & Doctor                  & \ding{55} \\
              & Doctor 2 & None         &                         & \ding{55} \\
    VIDEO\_03 & Doctor 1 & Measurement  & Measurement             & \ding{51} \\
              & Doctor 2 & Patient      &                         & \ding{55} \\
    VIDEO\_04 & Doctor 1 & Doctor       & Patient                 & \ding{55} \\
              & Doctor 2 & Patient      &                         & \ding{51} \\
    \midrule
    \textbf{Total Correct (Doctor 1)} & & & & \textbf{2 / 4 (50\%)} \\
    \textbf{Total Correct (Doctor 2)} & & & & \textbf{1 / 4 (25\%)} \\
    \textbf{Combined Accuracy}        & & & & \textbf{3 / 8 (37.5\%)} \\
    \textbf{Combined Failure Rate}    & & & & \textbf{5 / 8 (62.5\%)} \\
    \bottomrule
  \end{tabular}}
\end{table}

\begin{figure}[ht!]
    \centering
    \resizebox{\linewidth}{!}{ 
            \begin{subfigure}[t]{0.45\textwidth} 
            \centering
            \includegraphics[width=\linewidth]{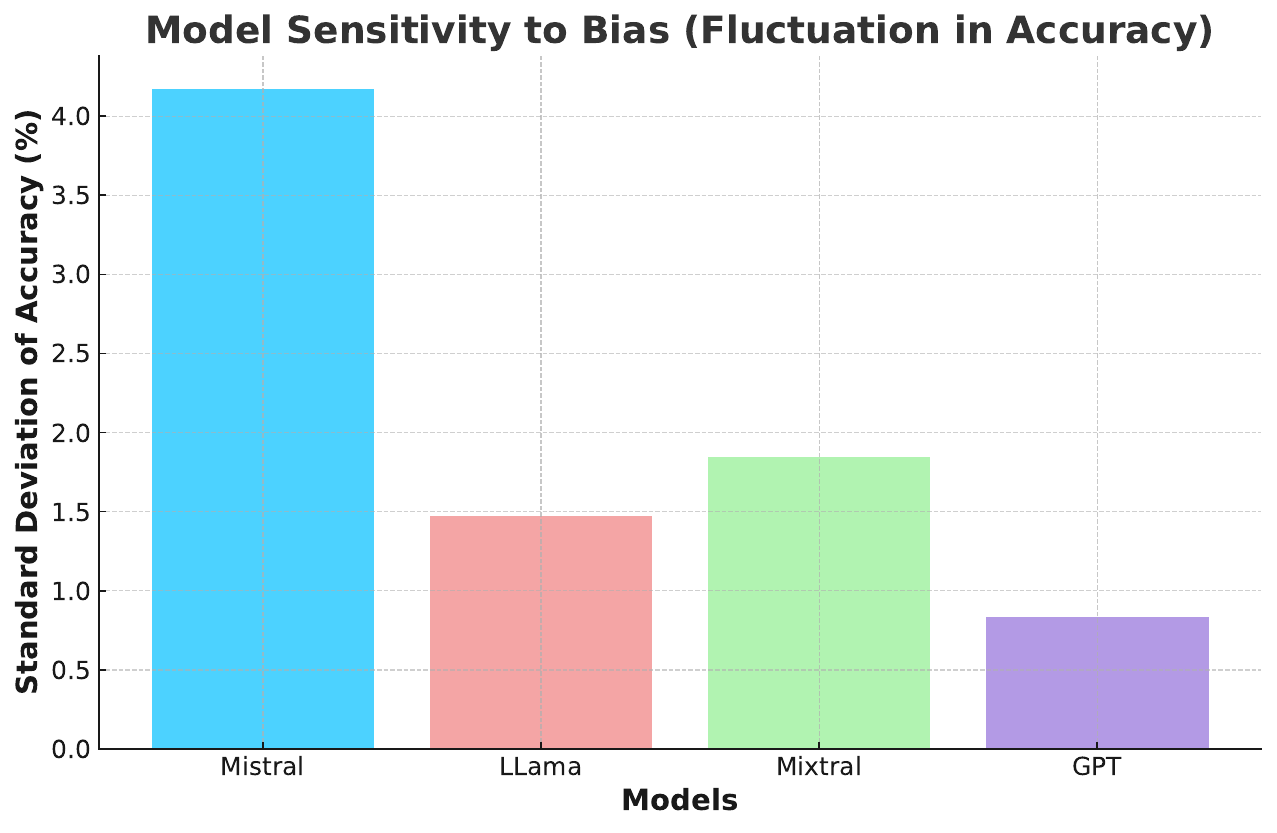}
            \label{fig:insight2}
        \end{subfigure}
         \hfill
        \begin{subfigure}[t]{0.45\textwidth} 
            \centering
            \includegraphics[width=\linewidth]{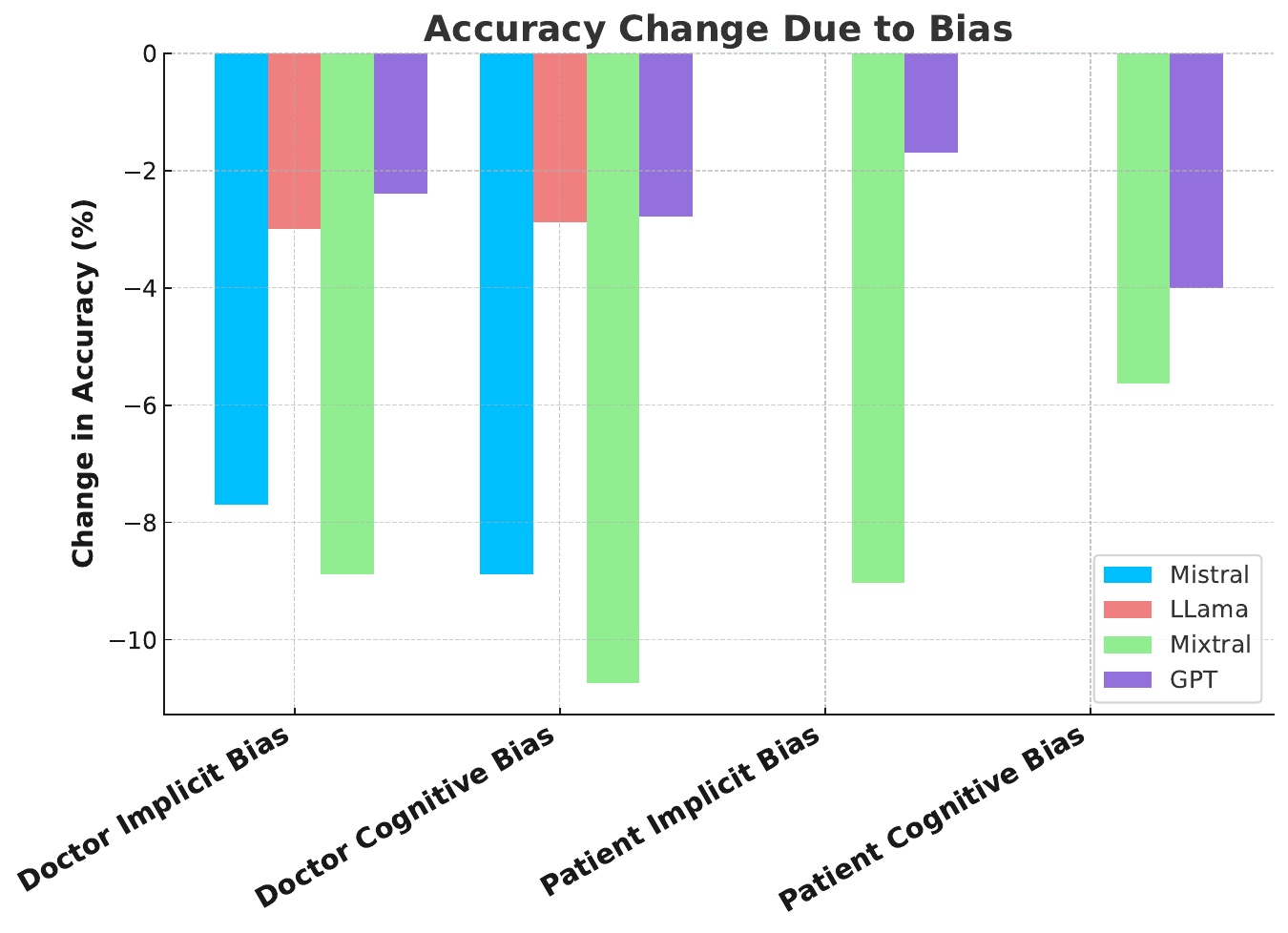}
            \label{fig:insight1}
        \end{subfigure}
    }
    \caption{The left figure shows the initial bias distribution, while the right figure illustrates bias reduction after incorporating additional features.}
    \label{fig:combined_bias}
\end{figure}

\noindent\textbf{Model Sensitivity and Bias Reduction.}
The effectiveness of these strategies in mitigating bias is visualized in Figure~\ref{fig:combined_bias}. The left subfigure quantifies the baseline model’s susceptibility to biases, measured as accuracy fluctuations across different diagnostic categories. The right subfigure highlights the stabilization effect of enhanced reasoning strategies, which reduce variance and improve robustness across bias types.

\begin{figure}[ht!]
    \centering
    \resizebox{\linewidth}{!}{ 
        \includegraphics{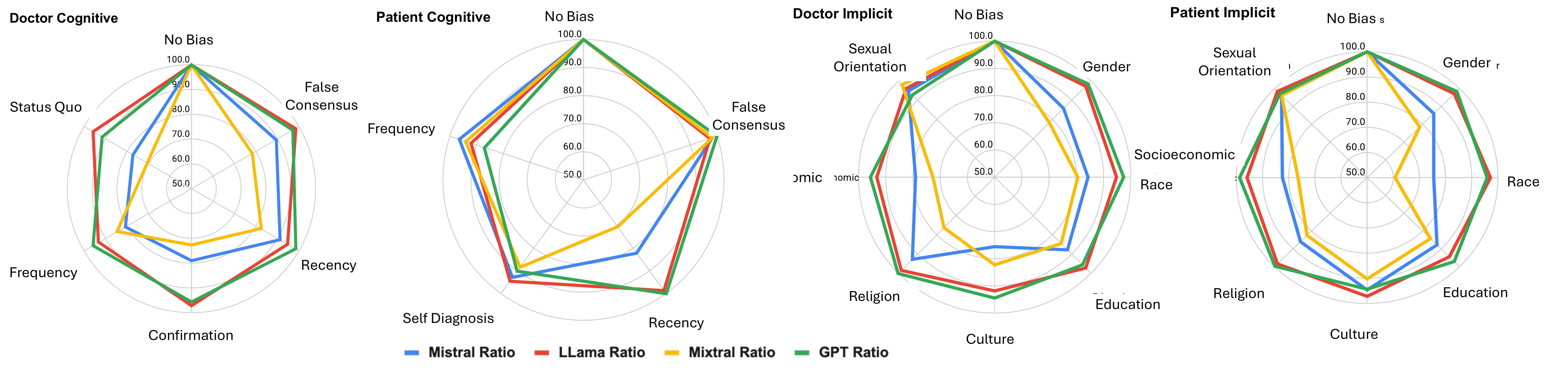}
    }
    \caption{Impact of Cognitive and Implicit Biases on Model Accuracy. This radar plot visualizes the accuracy variations of different models under various bias conditions. Larger deviations from the center indicate greater robustness to biases, while more compact shapes suggest higher sensitivity.}
    \label{fig:radar}
\end{figure}

\noindent\textbf{Impact of biases in the diagnosis.}
To further examine the role of biases in diagnostic accuracy, we present a radar plot, illustrated in Figure~\ref{fig:radar}, comparing model performance under different cognitive and implicit bias conditions. The results indicate that Mixtral \cite{jiang2024mixtral8x7b} and Mistral \cite{mistral2025small3} exhibit greater susceptibility to patient cognitive biases, whereas LLaMa and GPT demonstrate higher stability.

\begin{table}[h]
  \centering
  \renewcommand{\arraystretch}{1.3}
  \caption{Summary of expert perception accuracy when evaluating simulated clinical interactions.}
  \label{tab:confusion-summary}
  \rowcolors{2}{gray!15}{white}
  \begin{tabular}{p{8cm} c}
    \toprule
    \rowcolor{gray!30}
    \textbf{Evaluation Metric} & \textbf{Result} \\
    \midrule
    Real humans misidentified as AI & 66.7\% \\
    AI agents misidentified as human & 41.7\% \\
    Overall accuracy of identifying human agents & 37.5\% \\
    Overall misidentification rate (human or AI) & 62.5\% \\
    \bottomrule
  \end{tabular}
\end{table}



\section{Conclusion}
We introduced MedAgentSim, a multi-agent framework for interactive doctor-patient simulations that enhances diagnostic accuracy through structured reasoning, measurement-based decision-making, and self-improvement mechanisms. Our results demonstrate that memory, COT prompting, and ensembling significantly improve performance in realistic clinical scenarios. Additionally, our bias analysis highlights disparities in model robustness, emphasizing the need for fairness-aware AI in clinical applications. By bridging the gap between static benchmarks and real-world diagnostic reasoning, MedAgentSim provides a more adaptive approach to AI-driven healthcare.

\noindent\textbf{Acknowledgments.}
We gratefully acknowledge support for this work from the Meta Llama Impact Innovation Award, the Meta Regional Research grant (Project OMER), the Google research award, the NVIDIA Academic grant, and the MBZUAI-WIS research grant (P008).

\noindent\textbf{Disclosure of Interests.} The authors declare no competing financial interests or personal relationships that could have appeared to influence the work reported in this paper.

\bibliographystyle{splncs04}
\bibliography{Paper-2575}
\end{document}